\documentclass[sigconf,nonacm]{acmart}
\usepackage{float}
\usepackage{enumitem}
\usepackage{etoolbox}

\makeatletter
\pretocmd{\@mkabstract}{\vspace*{-0.025cm}}{}{}
\makeatother

\setcopyright{none}
\renewcommand\footnotetextcopyrightpermission[1]{}

\definecolor{newcontent}{HTML}{C92A2A}
\newif\ifnewcontentmarkers
\newcontentmarkerstrue

\newcommand{\filmstrip}[2][\linewidth]{%
  \IfFileExists{figures/#2_strip.jpg}%
    {\includegraphics[width=#1]{figures/#2_strip.jpg}}%
    {\fbox{\parbox[c][6em][c]{\dimexpr#1-2\fboxsep-2\fboxrule\relax}{\centering
      \sffamily\small\color{black!50}
      video filmstrip placeholder\\[2pt]\texttt{\detokenize{figures/#2_strip.jpg}}\\[2pt]
      (drop the source video in \texttt{assets/videos/} and run
      \texttt{make figures})}}}%
}
\newcommand{\chartimage}[2][\linewidth]{%
  \IfFileExists{#2}%
    {\includegraphics[width=#1]{#2}}%
    {\fbox{\parbox[c][10em][c]{\dimexpr#1-2\fboxsep-2\fboxrule\relax}{\centering
      \sffamily\small\color{black!50}
      chart placeholder\\[2pt]\texttt{\detokenize{#2}}\\[2pt]
      (export from the Notion page into \texttt{assets/images/})}}}%
}

\makeatletter
\patchcmd{\@mkauthors@iii}
  {\lineskip=1pc\relax}
  {\lineskip=0.85pc\relax}
  {}{}
\makeatother

\title{Solaris: Towards Interfaces That Are Generated, Not Coded}

\author{Yuval Alaluf}
\authornote{Core contributing team, listed in alphabetical order.}
\affiliation{\institution{Runway}\country{}}
\email{yuval@runwayml.com}

\author{Omri Avrahami}
\authornotemark[1]
\affiliation{\institution{Runway}\country{}}
\email{omri@runwayml.com}

\author{Guy Bukchin Leshem}
\authornotemark[1]
\affiliation{\institution{Runway}\country{}}
\email{guy@runwayml.com}

\author{Michal Geyer}
\authornotemark[1]
\affiliation{\institution{Runway}\country{}}
\email{michal@runwayml.com}

\author{Kfir Goldberg}
\authornotemark[1]
\affiliation{\institution{Runway}\country{}}
\email{kfir@runwayml.com}

\author{Elad Richardson}
\authornotemark[1]
\affiliation{\institution{Runway}\country{}}
\email{elad@runwayml.com}

\author{Diego Alarcón}
\affiliation{\institution{Runway}\country{}}
\email{diego@runwayml.com}

\author{Alejandro Alvarez}
\affiliation{\institution{Runway}\country{}}
\email{aalvarez@runwayml.com}

\author{Cole Garry}
\affiliation{\institution{Runway}\country{}}
\email{cgarry@runwayml.com}

\author{Anastasis Germanidis}
\affiliation{\institution{Runway}\country{}}
\email{anastasis@runwayml.com}

\author{Tenaya Goldsen}
\affiliation{\institution{Runway}\country{}}
\email{tenaya@runwayml.com}

\author{Corina Gurau}
\affiliation{\institution{Runway}\country{}}
\email{corina@runwayml.com}

\author{Robin Kahlow}
\affiliation{\institution{Runway}\country{}}
\email{robin@runwayml.com}

\author{Joel Kwartler}
\affiliation{\institution{Runway}\country{}}
\email{joel@runwayml.com}

\author{Kathleen Lewis}
\affiliation{\institution{Runway}\country{}}
\email{katie@runwayml.com}

\author{Alejandro Matamala Ortiz}
\affiliation{\institution{Runway}\country{}}
\email{alejandro@runwayml.com}

\author{Eugene McMahon}
\affiliation{\institution{Runway}\country{}}
\email{eugene@runwayml.com}

\author{Thon Prom}
\affiliation{\institution{Runway}\country{}}
\email{thon@runwayml.com}

\author{Sarah Saltonstall-Wurm}
\affiliation{\institution{Runway}\country{}}
\email{sarah@runwayml.com}

\author{Jamie Umpherson}
\affiliation{\institution{Runway}\country{}}
\email{jamie@runwayml.com}

\author{Hudson Yeo}
\affiliation{\institution{Runway}\country{}}
\email{hudson@runwayml.com}

\begin{document}

\begin{abstract}
Digital interfaces are traditionally implemented through intermediate representations such as code, requiring their appearance and behavior to be specified in advance. We introduce \emph{Solaris}{%
  \renewcommand{\thefootnote}{\textdagger}%
  \footnote{\url{https://runway.com/news/research/introducing-solaris}}%
  \addtocounter{footnote}{-1}%
}, an \emph{interface world model} that instead generates an interactive UI directly, frame by frame, in response to user actions. Solaris treats mouse interactions as conditioning signals and autoregressively synthesizes the resulting visual state at interactive speeds. To enable real-time generation while maintaining visual coherence over extended interactions, we combine autoregressive frame generation with few-step distillation and training on the model’s own outputs. A language model complements the visual world model by interpreting user intent and specifying how interactions should affect the generated environment, separating high-level reasoning from visual rendering. By generating both the appearance and behavior of an interface dynamically, Solaris enables open-ended interactions that need not be explicitly programmed in advance. We view interface world models as a step toward a new paradigm for software, where interfaces are generated and adapted continuously around user intent rather than implemented as fixed collections of predefined states and behaviors.

\begingroup
% \renewcommand{\thefootnote}{\textdagger}
% \footnotetext{%
%   \url{https://runway.com/news/research/introducing-solaris}%
% }
\endgroup

\end{abstract}

\maketitle

\vspace*{0.35cm}

% ---------------------------------------------------------------- introduction
\vspace{-0.15cm}
\section{Introduction}\label{sec:intro}

We introduce \emph{Solaris}, the first model in a new family of AI systems we call \emph{interface world models}. Solaris starts with a question: what happens when an operating system generates apps and websites as you use them?

Every operating system, from early terminals to Linux and macOS, has dictated what is rendered on screen and what happens when a person or program acts on it. Applications get built on top, and stay fixed until someone pushes an update. Solaris instead renders that layer directly. It is a real-time interactive model that generates the interface itself, frame by frame. Every frame is synthesized as you interact, allowing the interface to respond continuously to your actions.

Design is more visual than ever, with pixel-perfect mockups and image models that can generate entire screens that are nearly indistinguishable from finished products. But images do not run like a website or app. Every piece of software built today still requires a translation: the visual design must first be converted into an intermediate representation (e.g.\ code) before it can do anything.

\begin{figure*}[t]
  \centering
  % Source video frames: 90, 145, 154, 174, 205, 276, 303, 337
  % (Runway_Solaris_Ecomm-Fashion.mp4 @ 30 fps; see scripts/make_filmstrips.sh)
  \includegraphics[width=\textwidth]{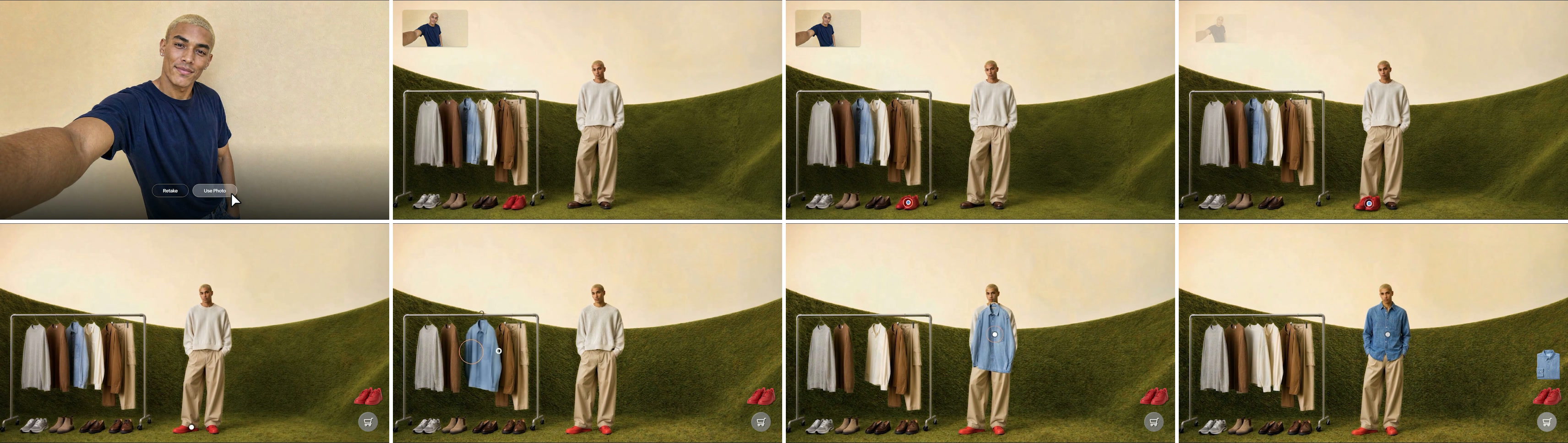} \\[-0.25cm]
    \caption{\emph{A virtual clothing store experience.} Solaris turns a visual environment into an interactive shopping experience: starting from a selfie, the user can try on products by dragging them directly onto themselves. Here, the user selects a pair of red shoes and then a blue shirt, with each item updating their appearance and being added to the cart. \\[-0.05cm]}
  \label{fig:ecomm}
\end{figure*}

\begin{figure*}[t]
  \centering
  % Source video frames: 36, 62, 78, 127, 171, 188, 239, 291
  % (Runway_Solaris_Interior.mp4 @ 24 fps; see scripts/make_filmstrips.sh)
  \includegraphics[width=\textwidth]{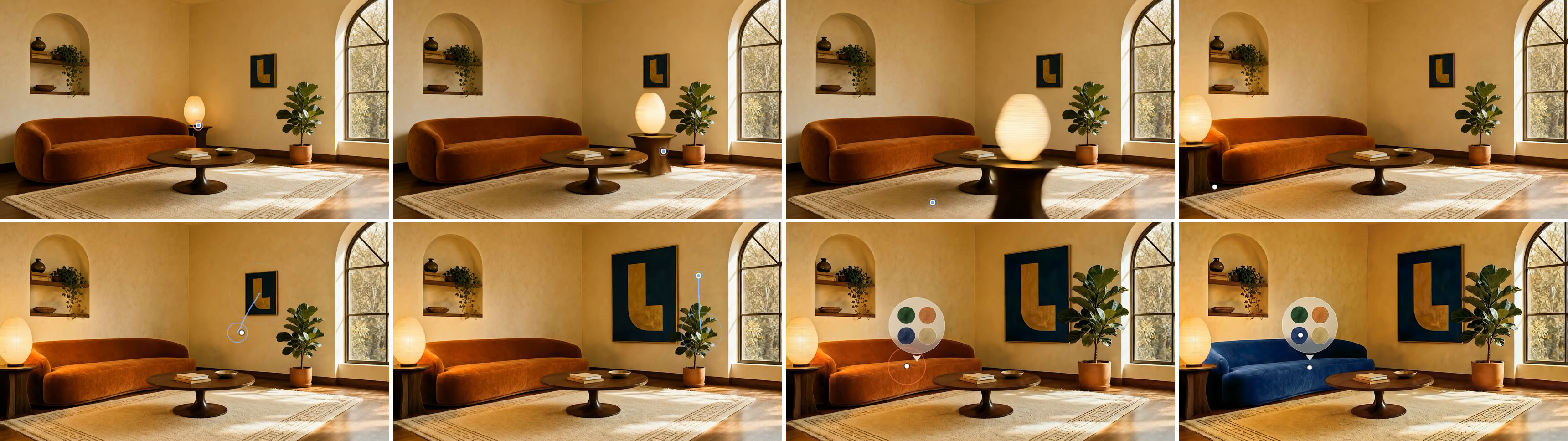} \\[-0.25cm]
    \caption{\emph{An interior room design experience.} The interface behaves like a living environment: the user can directly control objects and observe the scene respond in place. Here, the user moves a lamp, resizes the painting and plant, and applies a new fabric to the sofa. \\[-0.15cm]
    }
  \label{fig:interior}
\end{figure*}

That intermediate representation limits what an interface can be, and how it responds to human and agent interaction. Every behavior has to be explicitly defined and implemented ahead of time, so software ships as a lossy compression of the space of possible interactions, frozen before any user arrives. The same translation process also sacrifices visual fidelity. Once a design is reduced to a simplified representation, the interface can respond quickly, but only by giving up much of the richness of the original design.

Solaris handles rendering and interactions jointly, removing many of the tradeoffs we associate with design today. A single world model generates every frame and every response to user input, eliminating the need for an intermediate representation. Because there is no conversion step, there is no loss, and the entire frame becomes the interface.

We think Solaris opens up new ways of building websites, apps, and other online interfaces. But it is also a new way to train agents, in much more dynamic environments. Even the best LLMs today struggle to complete~\cite{osworld2} basic computer use tasks, like booking a hotel or ordering groceries. Because text-based models are being trained to use coded interfaces, they tend to learn the specific layout they were trained on, and cannot adapt to a slightly different interface (say, two different hotel websites). By collapsing the space between action and response, Solaris lets agents train against interfaces that are constantly changing, and layouts that may never have existed before.

% (the blog's "What's New" section is merged into the introduction)
Solaris brings three new capabilities to software.

\begin{figure*}[t]
  \centering
  % Top row source video frames: 0, 71, 160, 257
  % (Runway_Solaris_Open-Ended_X-Ray_1.mp4; see scripts/make_filmstrips.sh)
  \includegraphics[width=\textwidth]{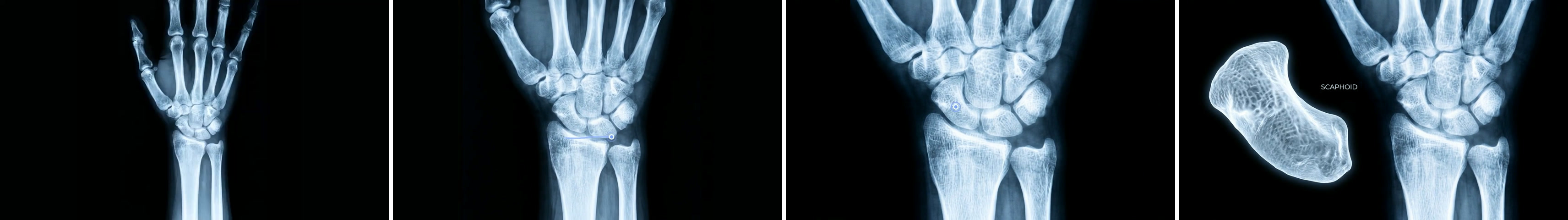}\\[-0.1cm]
  \textbf{Interaction $\#1$} \\[0.1cm]
  % Bottom row source video frames: 0, 36, 80, 150
  % (Runway_Solaris_Open-Ended_X-Ray_2.mp4; see scripts/make_filmstrips.sh)
  \includegraphics[width=\textwidth]{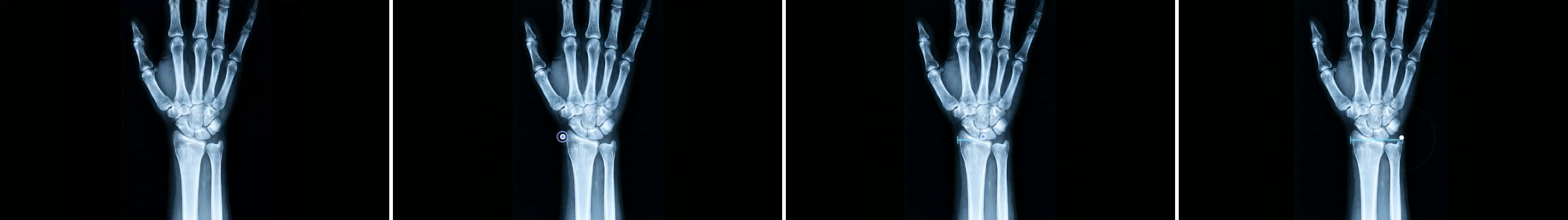}\\[-0.1cm]
  \textbf{Interaction $\#2$} \\[-0.25cm]
  \caption{\emph{Different interactions from the same initial state.} Interactions are not limited to a single predefined behavior: starting from the same frame, Solaris interprets the same drag interaction in different ways, either zooming in to inspect the wrist (top) or measuring the size of the hand (bottom).\\[0.05cm]}
  \label{fig:xray}
\end{figure*}

\begin{figure*}[t]
  \centering
  % Source video frames: 40, 66, 103, 113, 174, 210, 264, 411
  % (Nested_Sequence_09_prob4.mp4 @ 30 fps; see scripts/make_filmstrips.sh)
  \includegraphics[width=\textwidth]{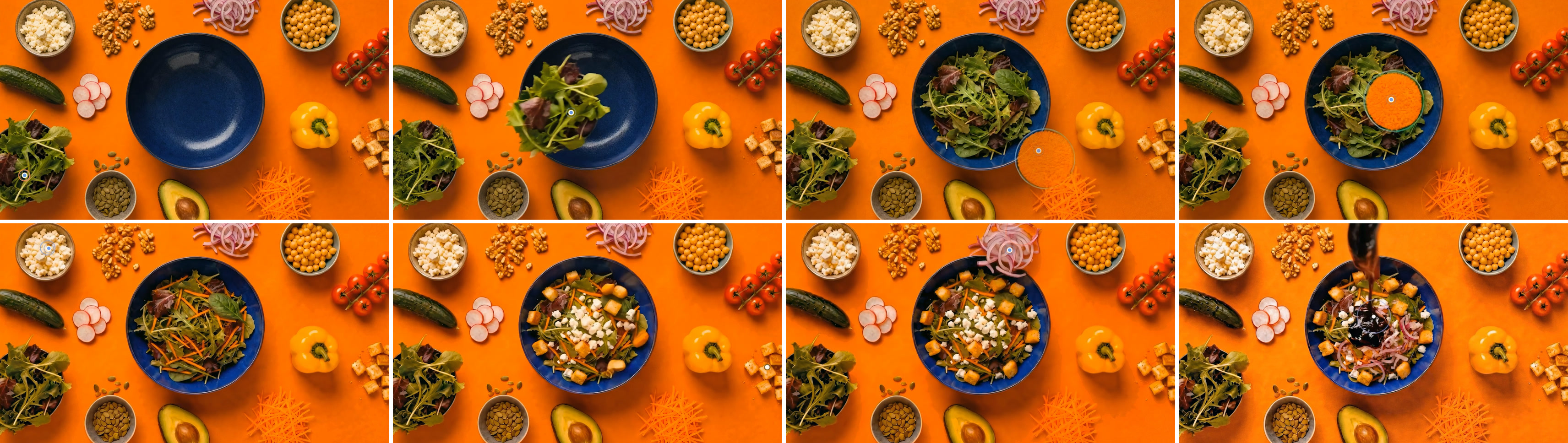}\\[-0.25cm]
  \caption{\emph{Building a salad.} 
  The user builds a salad directly in the scene by dragging ingredients into the bowl. Rather than selecting ingredients from a menu, they are arranged throughout the scene for direct interaction.
  \\[0.05cm]}
  \label{fig:salad}
\end{figure*}

\begin{figure*}[t]
  \centering
  % Source video frames: 0, 114, 138, 274, 356, 387, 606, 664
  % (Runway_Solaris_Edu-Fire.mp4 @ 30 fps; see scripts/make_filmstrips.sh)
  \includegraphics[width=\textwidth]{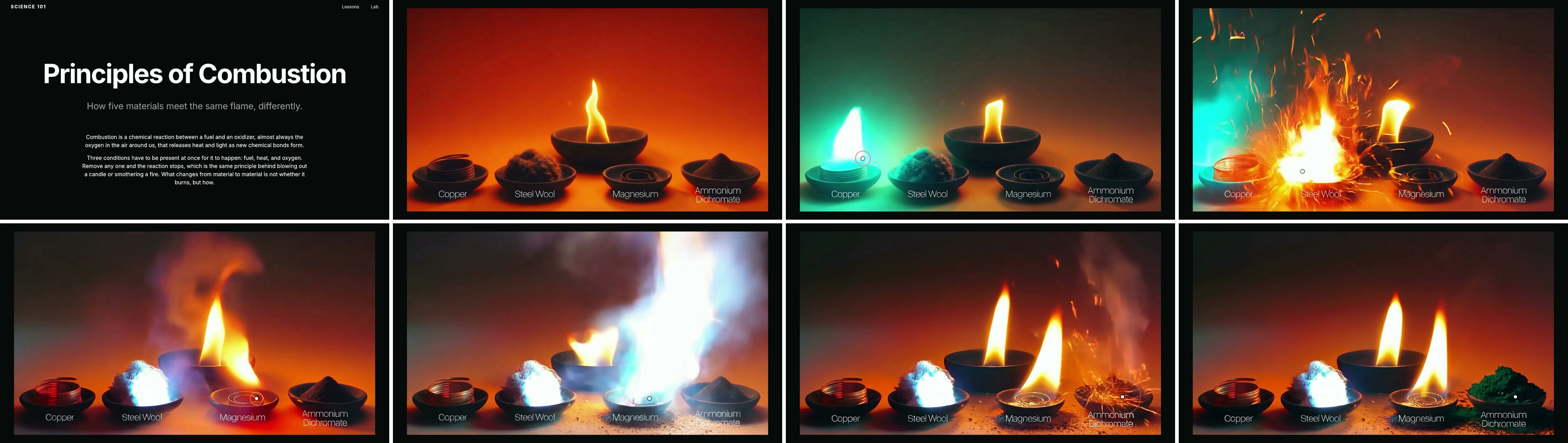}\\[-0.25cm]
  \caption{\emph{Interactive Principles of Combustion courseware.} Rather than presenting a static explanation, Solaris turns the lesson into an interactive experience. The user can experiment with different materials by dragging a flame onto them and directly observe how each reacts to fire.\\[0.05cm]}
  \label{fig:fire}
\end{figure*}

\begin{enumerate}[itemsep=0.25em]
    \item First, Solaris is entirely visual. When an image becomes the application
itself, there is no need for a second implementation step hidden beneath the
visuals that a user sees. Imagine browsing a virtual clothing store where the
showroom itself is the interface (Figure~\ref{fig:ecomm}). Using a single
image of yourself as a reference, you can pick up a shirt from a rack, drag it
onto yourself to try it on or rearrange the display as naturally as you would
in a physical store.

    \item Second, it is alive. Because the application is continuously rendered, it is
always evolving rather than waiting for the next user action
(Figure~\ref{fig:interior}). Reflections shift with the lighting, and objects
respond naturally as they are manipulated. A user can say something as simple
as: \emph{``Move the table so I can see how it looks'' or ``Change the color
of the couch.''} The result is software that feels less like navigating
through scripted pages and more like interacting with a living environment.

\begin{figure*}[t]
  \centering
  % Source video frames: 0, 38, 84, 129, 183, 348, 469, 570
  % (Runway_Solaris_Nav-Paris.mp4 @ 30 fps; see scripts/make_filmstrips.sh)
  \includegraphics[width=0.95\textwidth]{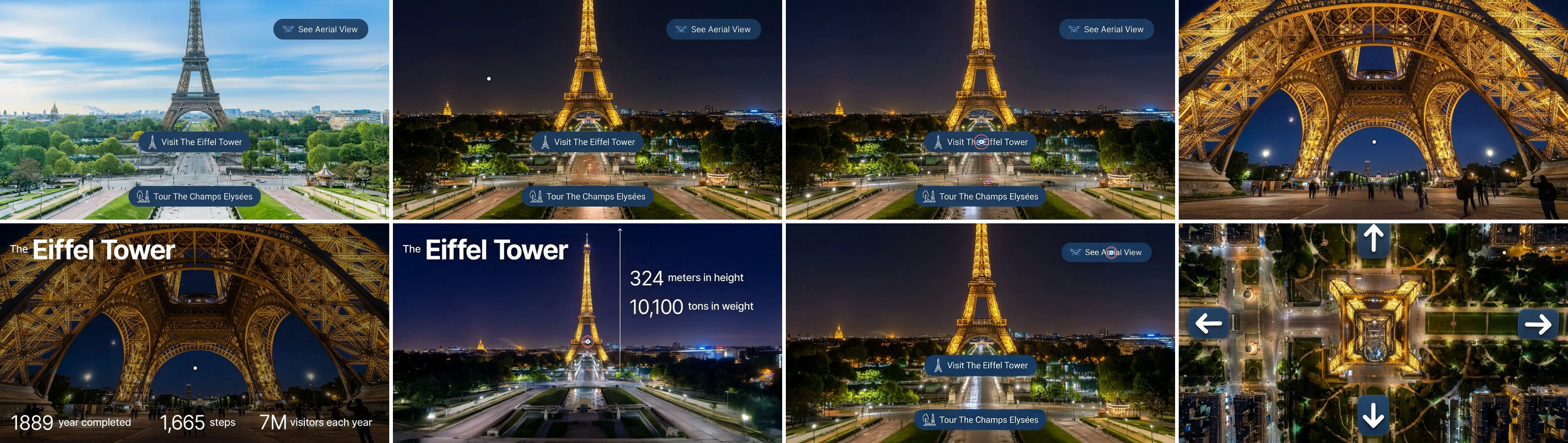}\\[-0.25cm]
  \caption{\emph{Navigating a real-world scene.} Rather than navigating between predefined views, the user explores the Eiffel Tower directly within the generated environment: changing the time of day, learning about the landmark, and viewing it from different perspectives while moving seamlessly through the scene.\\[-0.25cm]}
  \label{fig:paris}
\end{figure*}

\begin{figure*}[t]
  \centering
  % Source video frames: 39, 100, 165, 201, 361, 396, 507, 634
  % (Runway_Solaris_Alive_Car.mp4 @ 30 fps; see scripts/make_filmstrips.sh)
  \includegraphics[width=0.95\textwidth]{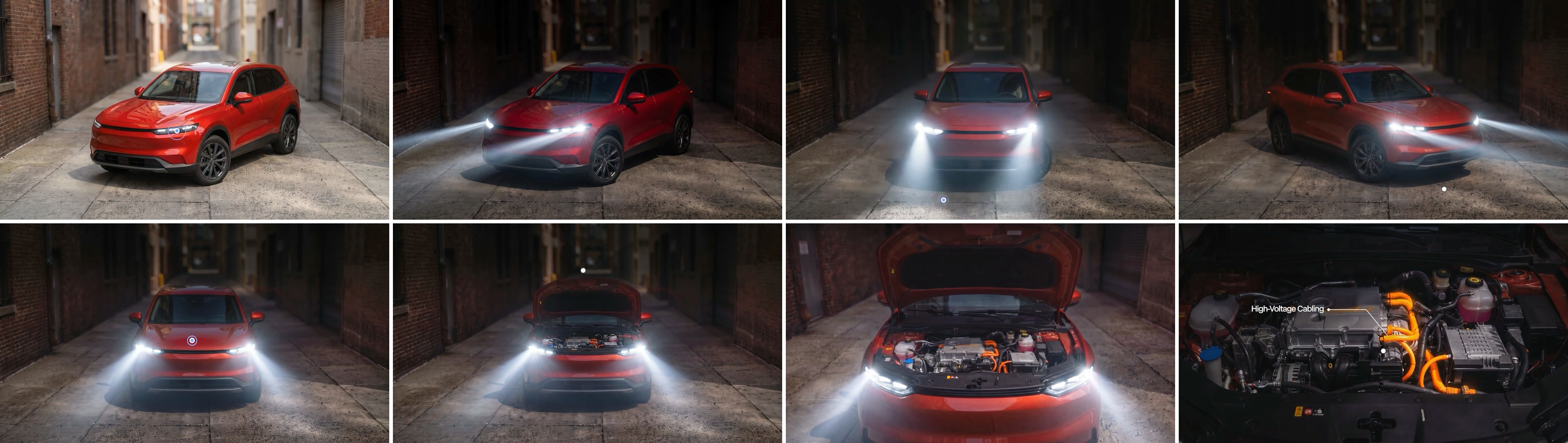}\\[-0.25cm]
  \caption{
  \emph{Exploring a product through direct interaction.} 
  The product itself becomes interactive: the user turns on the car's lights, drags the car to view it from different angles, and opens the hood to explore its internal components.
  \\[-0.25cm]}
  \label{fig:car}
\end{figure*}

    \item Finally, it is open-ended. Traditional interfaces are limited to the
interactions developers anticipated during development, but Solaris can
support entirely different behaviors in the same scene, reacting to user
interactions in real-time (Figure~\ref{fig:xray}). This flexibility decouples
the interface from predefined workflows, instead leaving the capabilities of
the driving world model to determine what is possible.
\end{enumerate}

Solaris turns an interface into an interactive experience rather than a
sequence of pages. Instead of selecting options from menus, users interact
directly with the scene itself. Building a salad is as simple as dragging
ingredients into a bowl, with the interface responding naturally as each
ingredient is added (Figure~\ref{fig:salad}).

Today, people turn to an LLM when they get stuck while online. But LLMs
answer in text, and most hands-on tasks are not text-based problems. Solaris
renders the next step in your context with visuals, and you can steer it. For
instance, you can generate an interactive demonstration of combustion,
allowing users to experiment with different materials and observe physically
plausible reactions as they interact (Figure~\ref{fig:fire}).

% ------------------------------------------------------------------- why now
\vspace*{-0.1cm}
\section[Motivation]{Motivation}\label{sec:motivation}

Digital interfaces are built on two systems, which until now have lived in
different worlds.
\begin{itemize}[itemsep=0.25em]
  \item The systems that know things (e.g., search engines and AI assistants)
    answer with static content: text, an image, maybe an embedded video.
  \item The systems that respond in real time (e.g., JavaScript/CSS, game engines and, more recently, interactive world models) create rich, interactive experiences, but they know nothing about your products or what you want to accomplish.
\end{itemize}

\begin{figure*}[t]
  \centering
  % Source video frames: 90, 145, 154, 174, 205, 276, 303, 337
  % (Runway_Solaris_Ecomm-Fashion.mp4 @ 30 fps; see scripts/make_filmstrips.sh)
  \includegraphics[width=\textwidth]{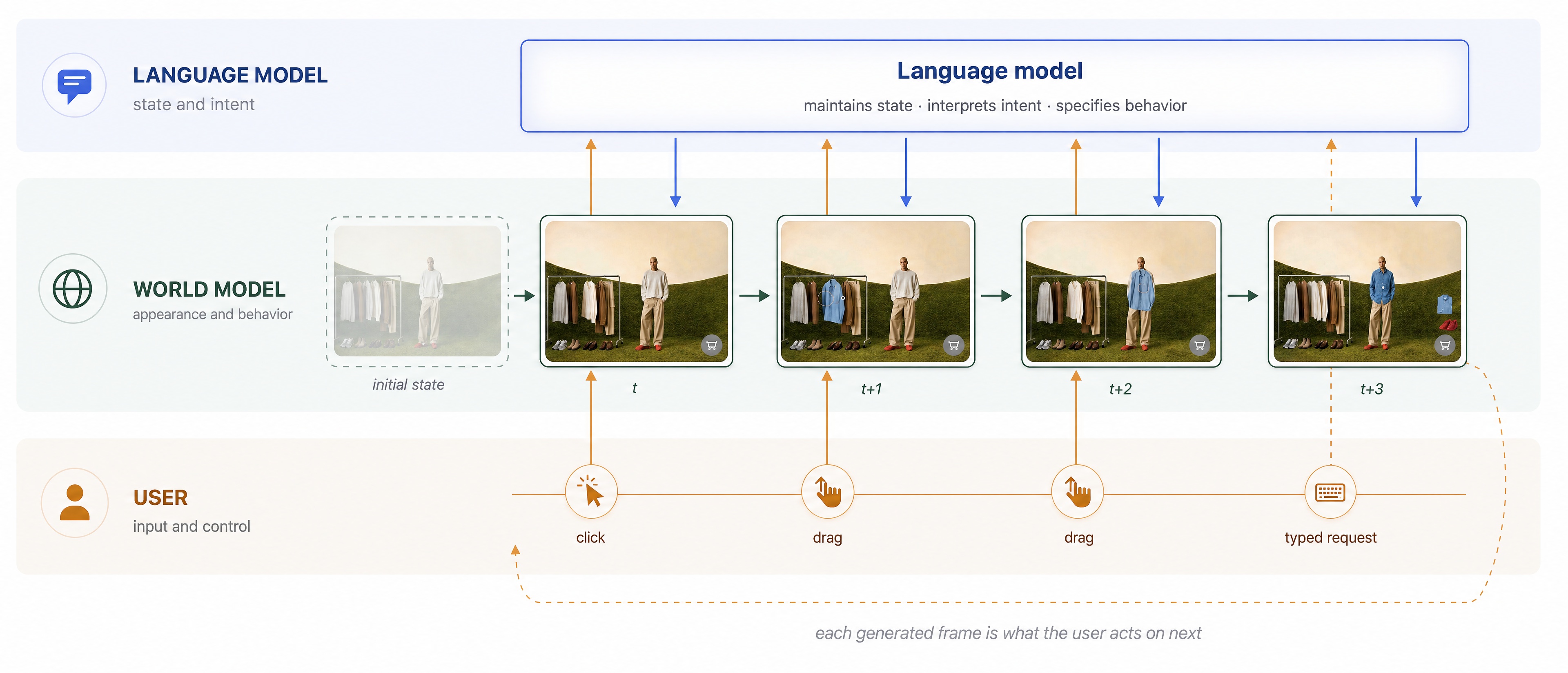} \\[-0.45cm]
\caption{\emph{How Solaris runs.} Solaris is organized as three levels. The user
acts on the frame with mouse and keyboard, the language model holds the state of
the session and specifies what each action should mean, and the world model
executes that intent, generating both the appearance and the behavior of the
result.\\[-0.05cm]}
  \label{fig:solaris_overview}
\end{figure*}

We have traditionally thought of software interfaces as deterministic programs
and world models as generators of visual content. An interface world model has
to be both at once: a system that understands your intent while continuously
rendering an interactive world around it.

Generative models have been applied to interfaces before, but primarily to reproduce software that already exists. NeuralOS~\cite{rivard2026neuralos} renders an Ubuntu desktop frame-by-frame from mouse and keyboard input, ViMo~\cite{luo2026vimo} generates the next screen of a mobile app as an image, and CUWM~\cite{guan2026computer} predicts the next state of a Microsoft Office application given a candidate action. In each case, the model is trained against a reference implementation, with the goal of reproducing its behavior. As a result, the interactions available to the user remain bounded by what the original software can do.
 
Solaris is different in that there is no reference implementation defining how the interface should behave. Instead, behaviors are determined by a language model as the session unfolds and rendered by a world model that has learned how both physical environments and interfaces respond to interaction. 
This allows the meaning of an interaction to depend on its context: the same drag can zoom into a wrist or measure a hand (Figure~\ref{fig:xray}), and interactions need not correspond to behaviors implemented in existing software. Rather than reproducing an existing interface, Solaris generates both the interface and its behavior as the user interacts with it.

Once you try to build such an interface world model, three engineering challenges appear immediately:
\begin{itemize}
  \item \textbf{Speed.} Interactions stop feeling interactive somewhere around
    half a second of delay. Video diffusion models take seconds or minutes to
    produce a clip, which is acceptable for content creation but too slow for
    an interface. To cross that threshold, the model has to generate frames
    sequentially, with each frame depending only on what came before, cheaply
    enough to keep up with the user.
  \item \textbf{Staying coherent.} An interface has to remain consistent
    across an entire session, not just a single clip. The things it needs to
    preserve (e.g., text, layout, the identity of objects) are the same things
    generated video has historically struggled to maintain, and small errors
    compound the longer generation continues.
  \item \textbf{Cost.} Generating every frame is still more expensive than
    serving a page that was built once. The same work that made Solaris real
    time also made it orders of magnitude cheaper to run than a standard video
    diffusion model, and the cost curve continues to improve.
\end{itemize}

Solaris is our bet that these conceptual and technical barriers can be
overcome. We built it with three focuses: real-time interaction, coherence
over an entire session, and visual quality at 720p.

\section[Method]{Method}\label{sec:method}

Solaris builds on our Gen-4.5~\cite{gen45} video generation model, which we
adapted to (1)~understand interaction and (2)~respond in real time. It follows
the path we opened with GWM-1~\cite{gwm1}, our general world model. We illustrate Solaris in Figure~\ref{fig:solaris_overview}.

\paragraph{\textbf{Learning interaction.}} Solaris treats user input as conditioning for
the next frame, the same way it treats text or images. The model observes
clicks, drags and other interactions as it generates, using them as signals
for what comes next. Because the model only ever sees interactions that have
already happened (never future ones), it learns the relationship between user
actions and visual outcomes. This means that it knows what should happen when
something is clicked, dragged or modified, without requiring those
interactions to be explicitly programmed.

\paragraph{\textbf{Running in real time.}} Standard video diffusion models refine an
entire clip over dozens of denoising steps, a process that is far too slow for
dynamic user interaction. We converted Solaris into a real-time engine in
three stages. First, we taught it to generate frames autoregressively, with
each frame depending only on what came before. Next, we distilled the
many-step denoising process into just a few steps. Finally, we trained the
fast model on its own outputs so visual quality remains stable over long
interactions. The result generates frames at interactive speeds while
preserving the visual quality of the original teacher model.

\paragraph{\textbf{Reasoning and rendering.}} Solaris generates the interface one frame
at a time, while a language model determines how that interface evolves. The
LLM interprets user requests, decides when interactions should modify the
current scene versus transition to a new one, defines the behaviors that make
the world feel alive and produces the prompts that guide Solaris as it renders
each state. Together, the language model and world model separate reasoning
from rendering: one decides what the application should do next, while the
other generates how that behavior appears and responds in real time.

\paragraph{\textbf{Continuous generation.}} You provide a starting state (e.g.\ a brand
environment or product scene) and the model streams frames in real time. As
the user clicks, drags or types, those interactions are incorporated into the
next generated frames, and the scene responds in place. There are no
predefined screens and no templates to fall back on. Instead, text prompts
specify what clicks, drags and other interactions mean in a particular scene.

\paragraph{\textbf{Redefining the mouse.}} Once interactions are described in natural
language instead of programmed, they no longer have to be fixed in advance.
Every object in the scene can become a new kind of tool. Click on a cat, and
your next clicks apply its fur color and texture to whatever you touch. Click
on a painting, and you might begin drawing in its style.

\section[Results]{Results}\label{sec:results}

\subsection{The Cost of Translation}\label{sec:cost-of-translation}

\begin{figure}[t]
  \centering
  \includegraphics[width=\columnwidth]{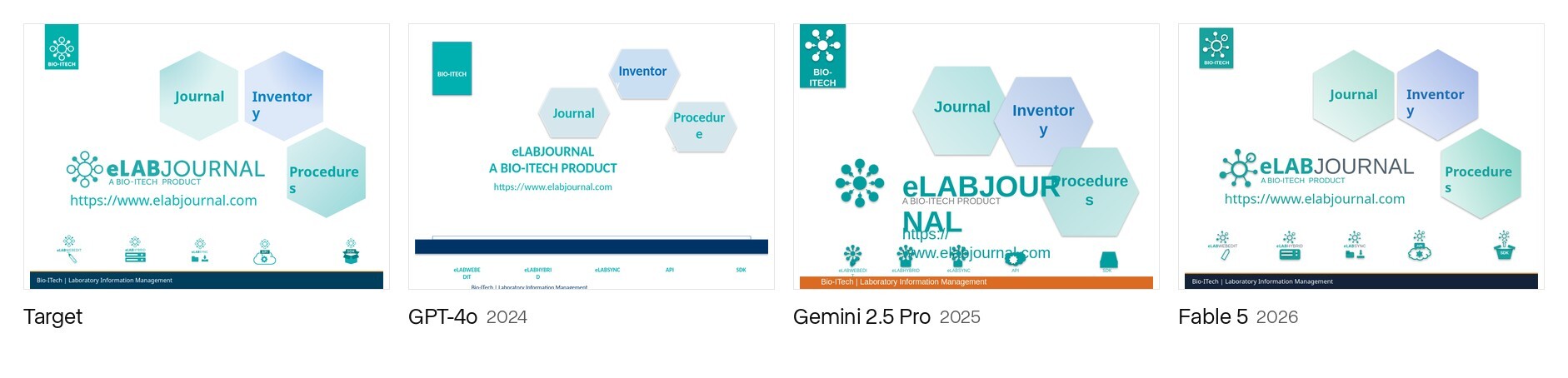}\\[1pt]
  \includegraphics[width=\columnwidth]{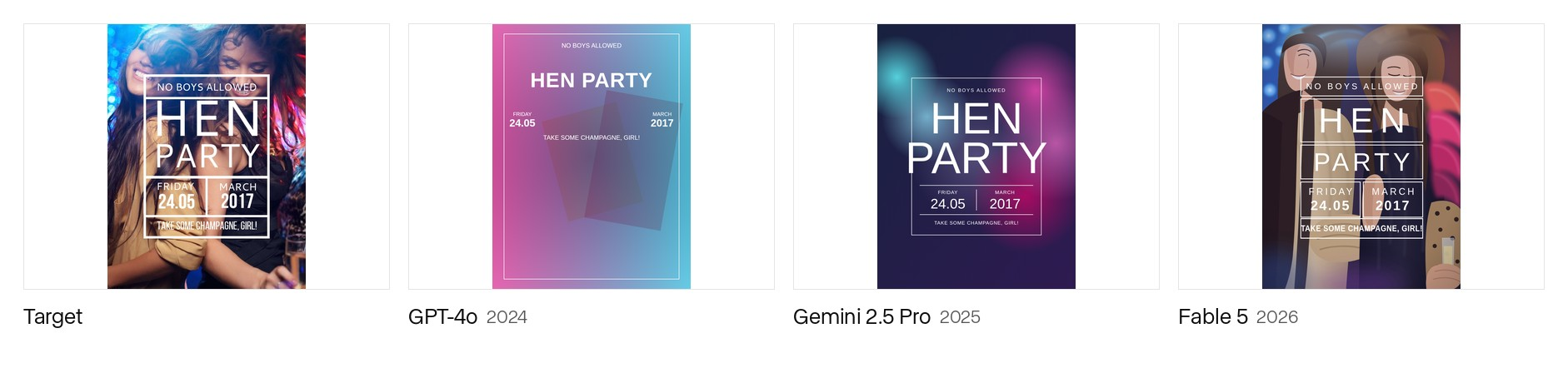}\\[1pt]
  \includegraphics[width=\columnwidth]{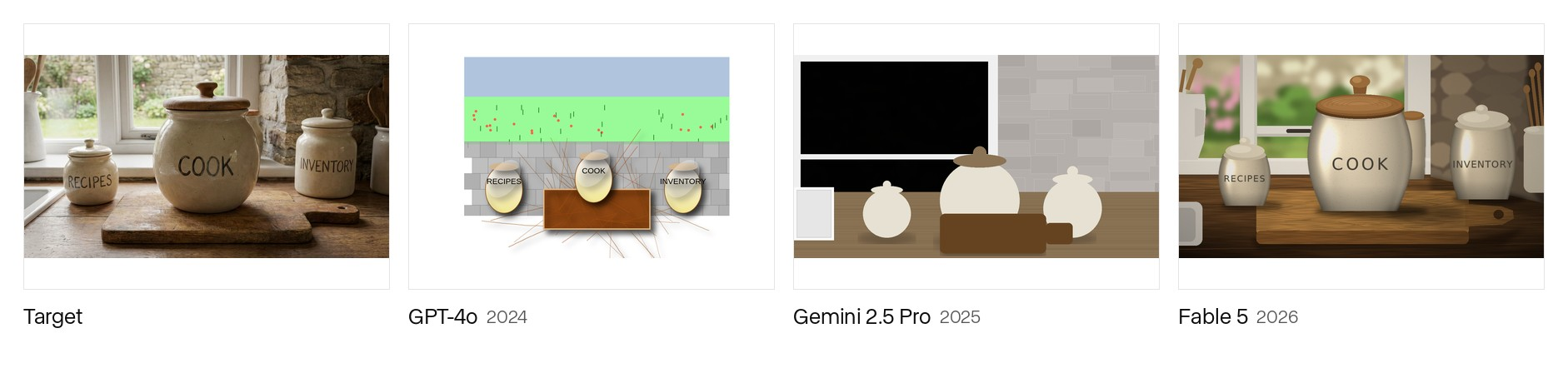}\\[-0.25cm]
  \caption{\emph{Reconstructing interfaces.}
    Given a single screenshot, state-of-the-art multimodal language
    models~\cite{gpt4o,gemini25,claude} attempt to reconstruct the visual
    interface through code. Reconstruction becomes increasingly difficult
    as visual complexity increases, from a plain webpage (top), to an
    image-heavy design (middle), to a natural image (bottom).}
  \label{fig:recon-examples}
\end{figure}

Earlier, we argued that translating interfaces into an intermediate
representation inevitably degrades information. 
To measure this, we evaluated state-of-the-art multimodal language models,
including Claude Fable 5~\cite{claude}, on the task of recreating website
interfaces from a single screenshot
(Figure~\ref{fig:recon-examples}). We evaluate across a diverse collection of
30 interfaces, ranging from simpler plain webpages to image-heavy webpages and
natural images, which evaluate different aspects of visual understanding.

We measure information preservation in two complementary ways. First,
structural similarity (SSIM)~\cite{ssim} compares the reconstructed
interface to the original in place, capturing how faithfully the visual
appearance is reproduced. Second, we compare each region of the original with
its most similar region anywhere in the reconstruction using
DINOv3~\cite{dinov3} features, measuring whether the underlying visual content
survives even when elements move or the layout changes.

% \begin{figure}[t]
%   \centering
%   \includegraphics[width=\columnwidth]{assets/images/slides_272_strip.jpg}\\[4pt]
%   \includegraphics[width=\columnwidth]{assets/images/recon_example_1.png}\\[4pt]
%   \includegraphics[width=\columnwidth]{assets/images/recon_example_2.png}
%   \caption{\new{\emph{Reconstructing interfaces from screenshots.}
%     Given a single screenshot, state-of-the-art multimodal language
%     models~\cite{gpt4o,gemini25,claude} attempt to reconstruct the visual
%     interface through code. Reconstruction becomes increasingly difficult
%     as visual complexity increases, from a plain webpage (top), to an
%     image-heavy design (middle), to a natural image (bottom).}}
%   \Description{Three rows comparing target interfaces with reconstructions by
%     GPT-4o, Gemini 2.5 Pro and Fable 5.}
%   \label{fig:recon-examples}
% \end{figure}

% \begin{figure*}[t]
%   \centering
%   \includegraphics[width=\textwidth]{assets/images/reconstruction_fidelity.png}
%   \caption{\textbf{Reconstruction fidelity across increasing visual
%     complexity.} \emph{Even as multimodal language models continue to improve,
%     reconstruction quality consistently degrades as visual complexity
%     increases, revealing the information lost when interfaces are translated
%     through language.}}
%   \Description{Line charts showing reconstruction quality dropping as visual
%     complexity increases.}
%   \label{fig:recon-fidelity}
% \end{figure*}

\begin{figure*}[t]
  \centering
  \includegraphics[width=0.925\textwidth]{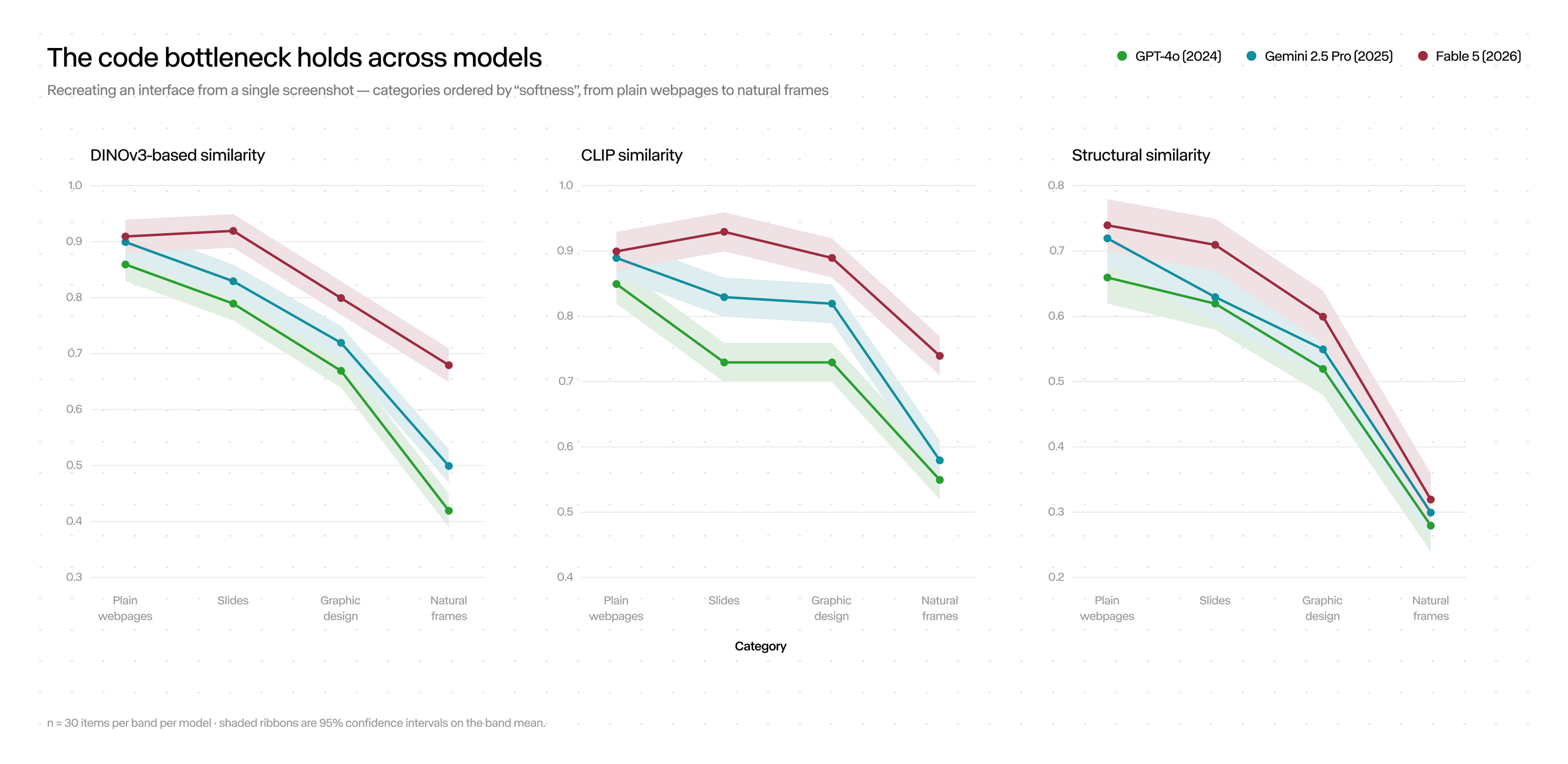}\\[-0.5cm]
  \caption{\emph{Reconstruction fidelity across increasing visual
    complexity.} Across multimodal language models, reconstruction
    fidelity decreases as visual complexity increases, highlighting the
    information lost when visual interfaces are translated through an
    intermediate representation.\\[-0.35cm]}
  \label{fig:recon-fidelity}
\end{figure*}

Despite rapid progress in recent years, every language model loses information
during reconstruction (Figure~\ref{fig:recon-fidelity}). Natural images are
affected most because rich visual detail cannot be represented accurately in
language. As interfaces become more complex, even small changes to text,
layout or structure can fundamentally alter how the interface behaves.

\begin{figure*}[t]
  \centering
  \begin{minipage}[t]{0.495\textwidth}
    \centering
    % Source video frames: 18, 69, 108
    % (Runway_Solaris_Alive_Mobile_Ocean.mp4 @ 30 fps)
    \includegraphics[width=\linewidth]{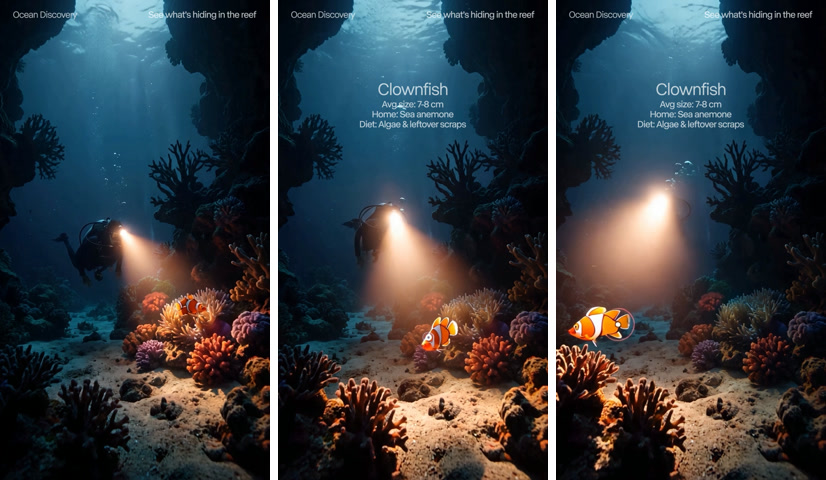}
  \end{minipage}\hfill
  \begin{minipage}[t]{0.495\textwidth}
    \centering
    % Source video frames: 59, 89, 134
    % (RW_INT_SOLARIS_SKATEBOARD_UI_UPSCALED.mp4 @ 30 fps)
    \includegraphics[width=\linewidth]{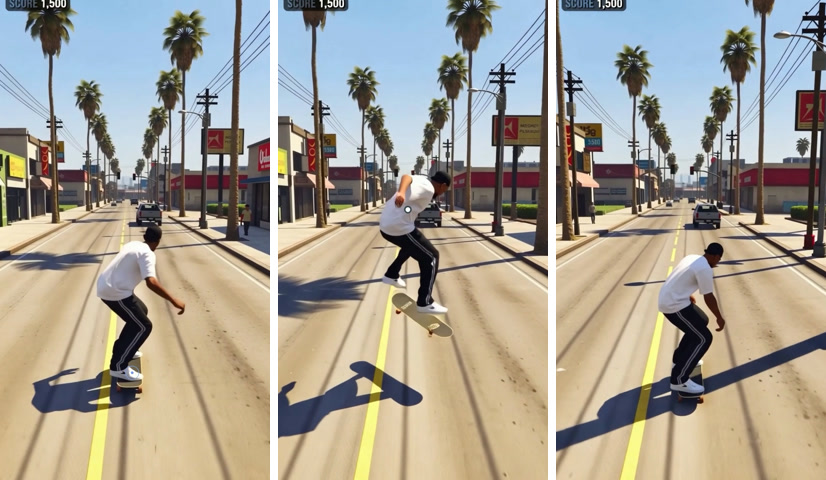}
  \end{minipage}\\[-0.2cm]
  \caption{\emph{Interactive mobile experiences.} Solaris turns generated mobile interfaces into responsive environments where user actions affect the scene coherently. The user drags a clownfish to a new location as the diver follows its movement (top), or drags a skateboard upward to make the rider jump (bottom). \\[0.05cm]}
  \label{fig:direct}
\end{figure*}

Rather than translating an interface into language and reconstructing it
again, Solaris operates directly on the visual interface itself. By
eliminating the intermediate representation, it preserves the complete visual
and semantic state of the interface from the very first frame
(Figure~\ref{fig:direct}).

% [H] (exact in-place) because a deferred single-column float queued behind
% the preceding figure*s is silently dropped at end of document.
\begin{figure}[t]
  \centering
  \includegraphics[width=0.95\columnwidth]{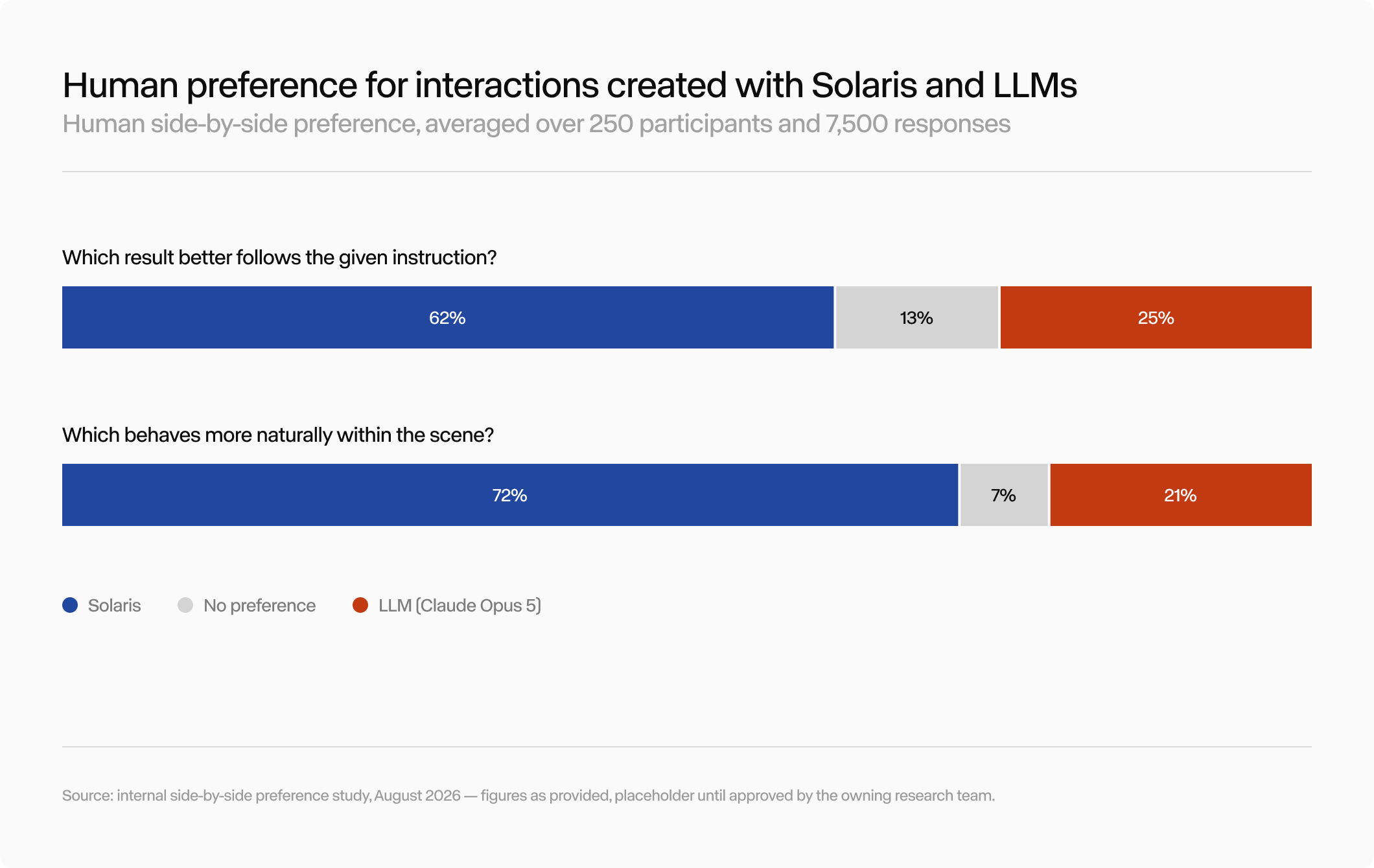}\\[-0.25cm]
  \caption{\emph{Interface world models produce more natural interactions.} Participants preferred Solaris over coded interfaces both for following the requested interaction and, by an even larger margin, for behaving naturally within the scene, highlighting the ability of interface world models to generate interactions that remain coherent with the environment.\\[-0.3cm]}
  \label{fig:user-study}
\end{figure}

\begin{figure*}[t]
  \centering

  % Top row
  \begin{minipage}[t]{0.495\textwidth}
    \centering
    {\sffamily\small\bfseries Solaris}\\[4pt]
    \includegraphics[width=\linewidth]{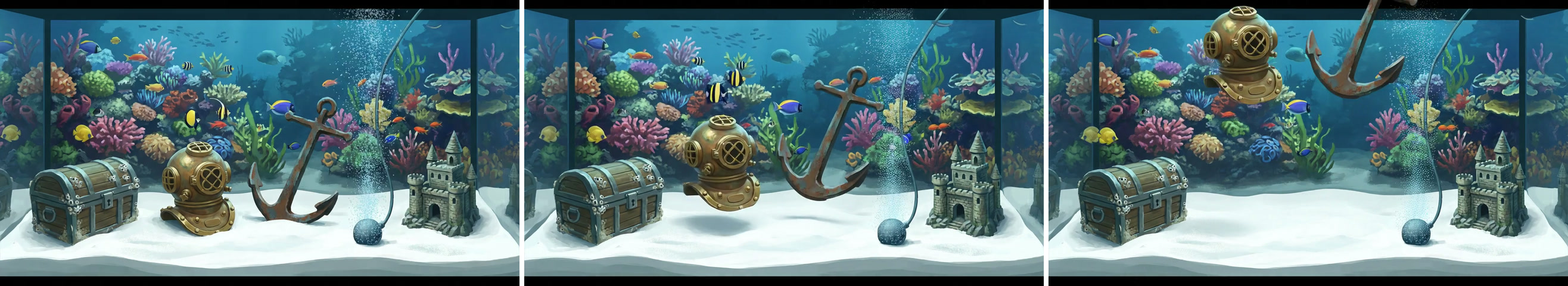}
  \end{minipage}\hfill
  \begin{minipage}[t]{0.495\textwidth}
    \centering
    {\sffamily\small\bfseries Coded interface (Claude Opus 5)}\\[4pt]
    \includegraphics[width=\linewidth]{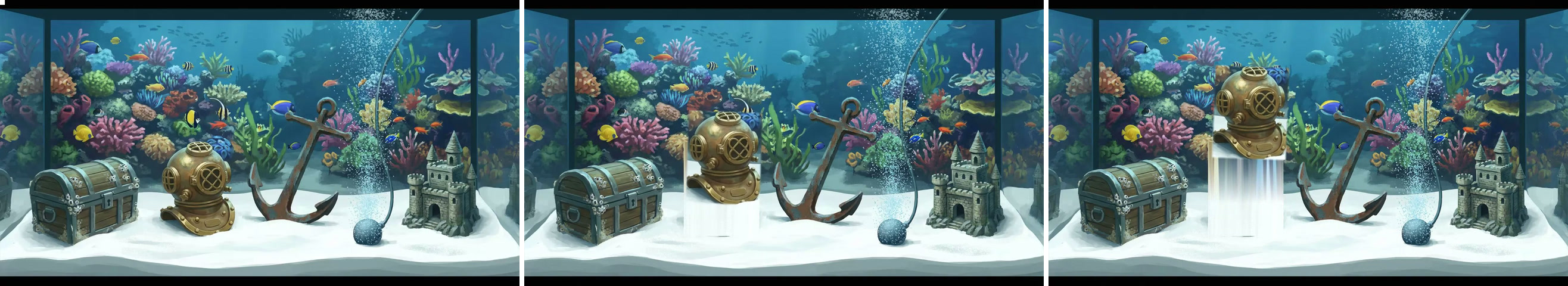}
  \end{minipage}

  \vspace{4pt}

  {\small\emph{``Clicking any element makes it lift off and float
  slowly upward, keeping its shape as it hovers.''}\par}

  \vspace{8pt}

  % Bottom row
  \begin{minipage}[t]{0.495\textwidth}
    \centering
    \includegraphics[width=\linewidth]{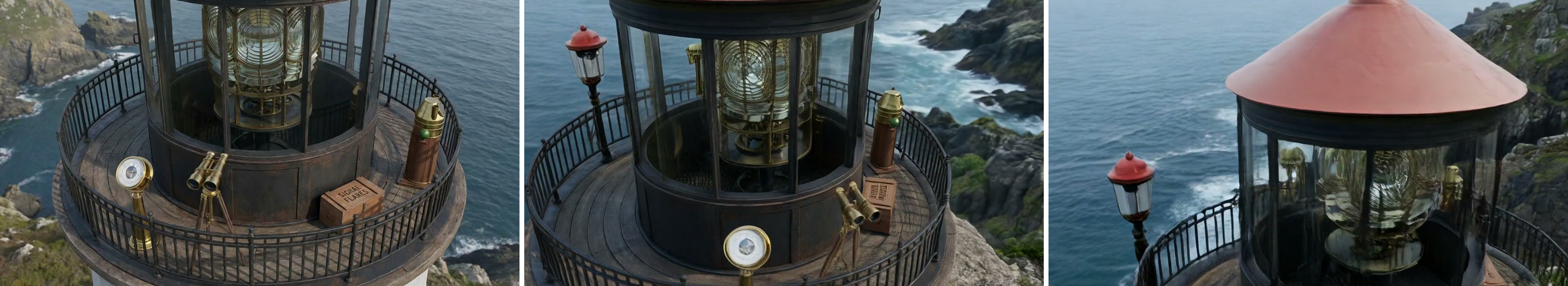}
  \end{minipage}\hfill
  \begin{minipage}[t]{0.495\textwidth}
    \centering
    \includegraphics[width=\linewidth]{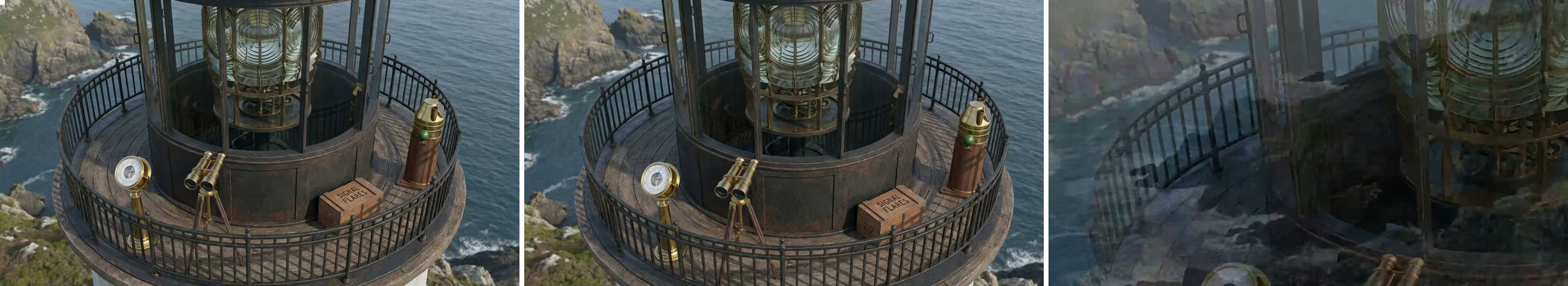}
  \end{minipage}

  \vspace{4pt}

  {\small\emph{``The camera pans right and tilts down in one
  smooth, continuous motion from the lighthouse balcony, settling on the
  waves crashing against the rocky coastline below.''}\par}

  \caption{\emph{Comparisons.} While both systems respond to the same
  interaction request, Solaris preserves the coherence of the entire scene,
  producing interactions that feel more natural and physically grounded.\\[0.15cm]}

  \label{fig:comparisons}
\end{figure*}

\subsection{Solaris vs.\ Coded Interfaces}
\label{sec:vs-coded}

Our reconstruction benchmark measures how much information is lost when an
interface is translated into code. We next ask: given the same interface and
the same user interaction, which approach produces the better result? Can a
coded interface recreate the same sense of a living, responsive environment as
an interface generated by an interface world model?

To answer this, we compared Solaris against a state-of-the-art language model
(Claude Opus 5~\cite{claude}). Both systems started from the same image and
received the same interaction requests, and we recorded how each responded
(Figure~\ref{fig:comparisons}). We then conducted a user study with 250
participants across 30 interaction examples, collecting nearly 7,500 pairwise
judgments. For each comparison, participants answered two questions:
``Which result better follows the given instruction?'' and ``Which behaves
more naturally within the scene?''

Participants preferred Solaris on both measures
(Figure~\ref{fig:user-study}). For following the requested interaction,
Solaris was preferred in 62\% of comparisons compared to 25\% for the coded
result, while 13\% were rated as equivalent. The difference was even larger
for natural behavior, where Solaris was preferred in 72\% of comparisons
compared to 21\% for the coded website, with 7\% rated as equivalent.

The second result highlights the broader difference between the two
approaches. A coded interface can often reproduce the requested change, but it
treats the interaction as an isolated update to the interface. With interface
world models, because the model already understands how objects, materials and
environments behave, it can generate interactions that feel coherent within
the scene rather than treating each UI action as an isolated element.

% ---------------------------------------------------------------- limitations
\vspace{-0.15cm}
\section[Limitations]{{Limitations}}\label{sec:limitations}

% (interior-design figure belongs to Section 7 but is placed early so the
% float lands on an earlier page instead of an orphaned trailing page)
\begin{figure*}[t]
  \centering
  % Source video frames: 9, 47, 129, 136, 198, 221, 265, 310
  % (interior_design.mp4 @ 24 fps; see scripts/make_filmstrips.sh)
  \includegraphics[width=\textwidth]{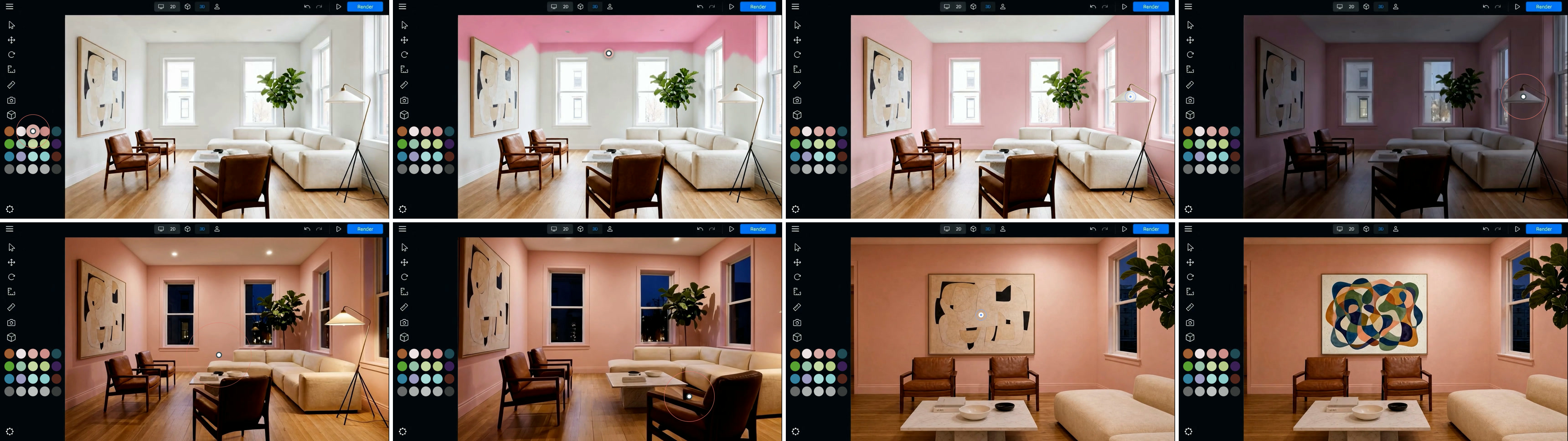}\\[-0.25cm]
  \caption{\emph{Interactive interior design.} A single generated interface supports a wide range of design interactions. The user can change wall colors, adjust the time of day and lighting, explore the room from different viewpoints, and modify artwork. \\[0.25cm]}
  \label{fig:interior-design}
\end{figure*}

Solaris is strongest at ambient motion, click-and-drag interactions and scene
transitions. Several important challenges remain:
\begin{itemize}[itemsep=0.3em, topsep=0.4em]
  \item \textbf{Text.} Stable, legible text remains one of the hardest
    problems in video generation, yet interfaces depend on it more than almost
    any other visual domain. One practical path is a hybrid system in which
    image models render text-heavy views whenever a brief pause is acceptable,
    while video models handle continuous interaction. Fully real-time
    generated text remains an open challenge.
  \item \textbf{Trust.} For instructional or commercial experiences, a
    convincing wrong answer is worse than no answer. Today, Solaris stays
    anchored through what you give it. The starting frame can be composed from
    real product imagery and reference material, which grounds the scene in
    things that actually exist. Conditioning generation on richer verified
    context as the session unfolds (reference images, product data, documents)
    is an active research focus.
  \item \textbf{Long sessions.} Maintaining visual and semantic coherence over
    extended, open-ended interactions remains an active area of research.
  \item \textbf{Accessibility and integration.} A generated interface still
    needs to work inside the rest of the software stack, including assistive
    technologies such as screen readers and accessibility APIs, so that
    flexibility does not come at the expense of usability.
\end{itemize}

These challenges reflect the current frontier of real-time generative models,
and we expect them to improve alongside the underlying models themselves.

% ------------------------------------------------------------ new interfaces
\vspace*{-0.15cm}
\section[Conclusions]{{Conclusions}}\label{sec:conclusions}

Solaris is an early step toward a new operating layer, and we see several new
interaction patterns emerging.
\begin{itemize}[itemsep=0.3em, topsep=0.4em]
  \item \textbf{The app stops being the unit you interact with.} Today, getting something done means opening the app made for it---one for shopping, another for news, another for restaurant reservations. If the operating system can generate useful interfaces, no matter what the user wants to do, there is less reason to sort software into a fixed catalog of apps. What you need simply shows up, customized to you.
    
  \item \textbf{Interface world models remove the need to translate between a
    visual idea and an intermediate representation.} Instead of working
    through UI frameworks, components, and code, any visual concept can become
    an interactive interface.
    
  \item \textbf{A storefront is no longer a fixed layout that every visitor
    sees.} It becomes a generated environment that preserves the brand's
    identity while adapting to each individual. Products, layouts, colors,
    materials and recommendations reshape around your intent in real time,
    allowing your brand and products to remain recognizable within
    hyper-personalized experiences.
    
  \item \textbf{Tutorials no longer replay the same sequence for everyone.}
    Instead, they render the next step in your own context, adapt as you make
    progress and recover naturally when you go off script.
\end{itemize}

We expect interface generation to follow the same trajectory as image and
video generation: every model generation will become faster, more coherent,
more controllable and more capable. The challenges that once made generated
interfaces seem impractical now look increasingly like solvable engineering
problems.

Solaris is our first interface world model, and we are excited to continue
exploring what generated software can become, from richer interactions,
stronger grounding and longer-lived experiences to entirely new kinds of
interfaces that do not exist today. We are working with key partners to launch
Solaris publicly.

\newpage

% ---------------------------------------------------------------- references
\bibliographystyle{ACM-Reference-Format}
\bibliography{references}

\end{document}